\pdfoutput=1

\documentclass[11pt]{article}

\usepackage[]{acl}

\usepackage{times}
\usepackage{latexsym}
\usepackage{booktabs}
\usepackage{amsmath}
\usepackage{multirow}
\usepackage{hyperref}
\usepackage[normalem]{ulem}
\useunder{\uline}{\ul}{}
\usepackage[pdftex]{graphicx}
\usepackage{tabularx}
\usepackage{xspace}

\newcolumntype{L}{>{\scriptsize}l}

\usepackage[T1]{fontenc}

\usepackage[utf8]{inputenc}

\usepackage{microtype}

\usepackage{inconsolata}

\newcommand{\draftonly}[1]{#1}
\renewcommand{\draftonly}[1]{}

\newcommand{\toward}{\textsc{Toward}{}\xspace}
\newcommand{\away}{\textsc{Away}{}\xspace}
\newcommand{\name}{\textsc{Astrapop}{}\xspace}
\newcommand{\longname}{\textsc{A}uthorship \textsc{S}tyle \textsc{TRA}nsfer with \textsc{P}olicy \textsc{OP}timization{}\xspace}

\title{Authorship Style Transfer with Policy Optimization}

\author{Shuai Liu \and Shantanu Agarwal \and Jonathan May \\
        Information Sciences Institute \\ University of Southern California \\ \texttt{\{liushuai, shantanu, jonmay\}@isi.edu}}

\begin{document}
\maketitle

\begin{abstract}

Authorship style transfer aims to rewrite a given text into a specified target while preserving the original meaning in the source.
Existing approaches rely on the availability of a large number of target style exemplars for model training. 
However, these overlook cases where a limited number of target style examples are available.
The development of parameter-efficient transfer learning techniques and policy optimization (PO) approaches suggest lightweight PO is a feasible approach to low-resource style transfer.
In this work, we propose a simple two-stage tune-and-optimize technique for low-resource textual style transfer.
We apply our technique to authorship transfer as well as a larger-data native language style task and in both cases find it outperforms state-of-the-art baseline models.\footnote{Code and models sufficient for a reproducibility study are available at \url{https://github.com/isi-nlp/ASTRAPOP}.}

\end{abstract}
\section{Introduction}

Given a text authored by an arbitrary source author, can we make it look like it is written by an arbitrary target author without changing its meaning?
This is the domain of authorship style transfer.
In the era of large language models (LLMs), the promise of authorship style transfer can turn any LLM into our own personalized model by transferring the outputs into our own style, and also prevent our text from being identified by authorship identification models through transferring our texts into the style of another author.
This task is first studied as a classic text style transfer task that requires a large number of texts in the target style to develop the transfer model, which limits its application to only famous authors like Shakespeare \citep{xu-etal-2012-paraphrasing, krishna-etal-2020-reformulating}.

Recently, \citet{patel2023lowresource} propose a more general and practical task, low-resource authorship style transfer which can apply to non-famous authors who only have a limited number of texts.
To solve this new task, they develop an LLM-based approach, STYLL which transfers a text by prompting LLMs with several texts written by the target author.
Though intended to be a simple baseline, STYLL proves remarkably adept at style alteration. Deeper investigation by \citet{patel2023lowresource} shows that while the alteration does manage to remove, or move away from the original author's style, it is rather unable to adopt, or move toward, the intended target author.

STYLL is an entirely in-context learning (ICL) method; it uses no model training or modification. This is justified by \citet{patel2023lowresource} as, due to small amounts of style-relevant training data, methods that use supervised fine-tuning (SFT) such as STRAP \cite{krishna-etal-2020-reformulating} do not outperform ICL. In this work we instead consider whether this limited data can be repurposed, specifically as training for a \textit{style critic model}, thereby enabling a policy optimization (PO) approach to directly encourage text generation in the desired style.
Rather than train the model on pseudo-parallel data with the language modeling loss, we could use policy optimization (PO) approaches to directly optimize the model to maximize the authorship style transfer objective.

\begin{figure*}[t!]
    \centering
    \includegraphics[width=2.0\columnwidth]{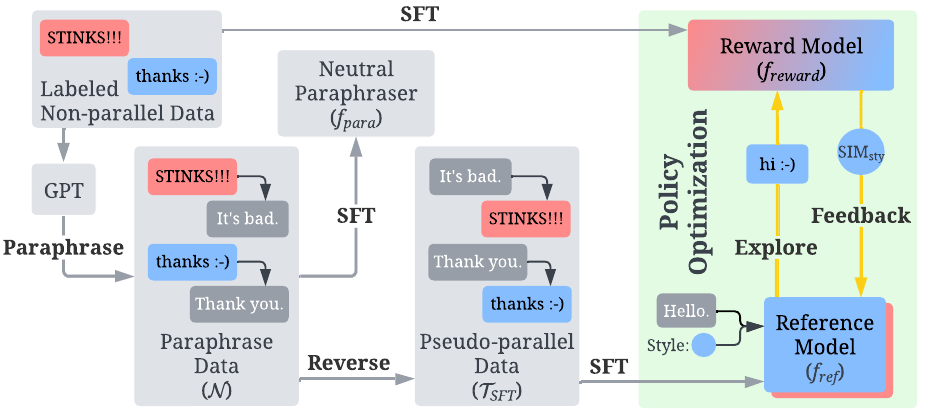}
    \caption{Overview of \name.}
    \label{fig:po_overview}
    \vspace{-0.4cm}
\end{figure*}

In this work, we propose \longname (\name), a lightweight two-stage PO training framework for authorship style transfer, in which the first supervised fine-tuning stage prepares a reference model for the second stage, and the policy optimization in the second stage further improves the performance of the reference model by directly optimizing it on the authorship style transfer objective.
Unlike more complicated RL-based approaches like \citet{hallinan-etal-2023-steer}, \name is more computationally efficient and flexible, and works well with a variety of both RL-based and RL-free PO algorithms.
We evaluate \name on two authorship style transfer tasks, a low-resource individual authorship style transfer task\footnote{with as few as five examples per author.} and a medium-resource community authorship style transfer task.
The evaluation results show that \name is more effectively able to leverage few-shot style transfer than ICL or SFT methods alone on the former task and also outperforms the state-of-the-art style transfer model with much less training time on the latter.

\section{Methodology}

In this section we formalize the authorship style transfer task and introduce \name.

\subsection{Task Definition}

The goal of authorship style transfer is to modify the style of the input text to make it look like the style of another author.
Formally, we have a dataset of texts with authorship style labels $\mathcal{D} = \{(\textbf{x}_1, s_1), (\textbf{x}_2, s_2), \cdots, (\textbf{x}_n, s_n)\}$ where the style label could be either at the individual level or the community level.
For convenience, we denote the semantic similarity between two texts $\textbf{x}_i$ and $\textbf{x}_j$ as $SIM_{sem}(\textbf{x}_i, \textbf{x}_j)$, and the similarity between the style of a text $\textbf{x}$ and a style $s$ as $SIM_{sty}(\textbf{x}, $s$)$.
Given an input text $\textbf{x}_s$ with style $s$ and a target style $t$, an authorship style transfer model rewrites $\textbf{x}_s$ into a new text $\textbf{x}_{s\rightarrow t}$ that maximizes $SIM_{sem}(\textbf{x}_{s\rightarrow t}, \textbf{x}_s)$ and $SIM_{sty}(\textbf{x}_{s\rightarrow t}, t)$, and minimizes $SIM_{sty}(\textbf{x}_{s\rightarrow t}, s)$.
We refer to maximizing $SIM_{sty}(\textbf{x}_{s\rightarrow t}, t)$ and minimizing $SIM_{sty}(\textbf{x}_{s\rightarrow t}, s)$ as the \textit{\toward} and \textit{\away} objectives, respectively.

\subsection{Framework Overview}
\name contains two main stages: \textbf{supervised fine-tuning} and \textbf{policy optimization}.
The framework overview is shown in \autoref{fig:po_overview}.
In the supervised fine-tuning stage, we train a reward model on labeled non-parallel data and a reference model on parallel in-domain data for policy optimization.
Due to a lack of parallel authorship style transfer data, we use the style transfer via paraphrasing (STRAP) strategy described in \citet{krishna-etal-2020-reformulating} to generate pseudo-parallel data to train the reference model.
Then, in the policy optimization stage, we directly optimize the reference model from the SFT stage on the \toward and \away objectives.

\subsection{Supervised Fine-tuning}
\label{sec:method_sft}

In this stage, we train three models with supervised fine-tuning: a \textbf{neutral paraphraser} $f_{para}$, a \textbf{reference model} $f_{ref}$, and a \textbf{reward model} $f_{reward}$.
$f_{para}$ is used for inference only, while $f_{ref}$ and $f_{reward}$ are used for PO training.

\subsubsection{Data Generation}

We first generate the pseudo-parallel training data for the neutral paraphraser and the reference model.
Following \citet{krishna-etal-2020-reformulating}, we generate neutral-to-target style transfer pairs by paraphrasing the target style texts with a neutral paraphraser.
To ensure the quality of the training data, we generate the neutral paraphrases with GPT-3.5-turbo using the same paraphrase prompt as in \citet{patel2023lowresource}.
Concretely, we generate neutral paraphrases for all texts in the dataset $\mathcal{D}$ to obtain a new set of texts $\mathcal{P} = \{\textbf{y}_1, \textbf{y}_2, \cdots, \textbf{y}_n\}$ where $\textbf{y}_i\in\mathcal{P}$ is the neutral paraphrase of $\textbf{x}_i\in\mathcal{D}$.
Then, we can build a neutral paraphrase dataset $\mathcal{N}=\{(\textbf{x}_1\rightarrow \textbf{y}_1), \cdots, (\textbf{x}_n\rightarrow \textbf{y}_n)\}$ and a neutral-to-target style transfer dataset $\mathcal{T}_{SFT}=\{(\textbf{y}_1\rightarrow \textbf{x}_1, s_1), \cdots, (\textbf{y}_n\rightarrow \textbf{x}_n, s_n)\}$.

\subsubsection{Paraphraser \& Reference Model}
We fine-tune off-the-shelf language models on the two generated supervised datasets to build the neutral paraphraser\footnote{Please see \autoref{sec:paraphraser_selection} for why we choose to train a local model instead of using GPT as the paraphraser.} and the reference model. 
Specifically, for the neutral paraphraser, we simply fine-tune the model on dataset $\mathcal{N}$ to maximize
\begin{equation*}\label{sft-par-obj}
    p(\textbf{y}|\textbf{x}) =
    \prod_{i=1}^{|\textbf{y}|}
    p(y^i | \textbf{x}, \textbf{y}^{<i})
\end{equation*}
where $\textbf{y}^{<i}$ represents tokens preceding token $y^i$ in $\textbf{y}$.
Similarly, for the reference model, we fine-tune the model on dataset $\mathcal{T}_{SFT}$ to maximize
\begin{equation*}\label{sft-trf-obj}
    p(\textbf{x}|\textbf{y}, s) =
    \prod_{i=1}^{|\textbf{x}|}
    p(x^i | \textbf{y}, \textbf{x}^{<i}, s)
\end{equation*}
Note that the probability in the training objective for authorship style transfer is additionally conditioned on the target style $s$.
Following \citet{wolf2019transfertransfo}, we implement all seq2seq models using decoder-only transformers.

\subsubsection{Reward Model}

Besides the reference model, PO training also requires a reward model to measure the style similarity $SIM_{sty}$.
We train a style model on dataset $\mathcal{D}$ to serve this purpose.
The details on how to train the style model and calculate $SIM_{sty}$ from the style model output are shown in \autoref{Reddit_Reward_Model} and \autoref{ETS_Reward_Model}.

\subsection{Policy Optimization}
We further train the reference model $f_{ref}$ using policy optimization and the reward model $f_{reward}$ to obtain the final \textbf{PO transfer model} $f_{PO}$.

\subsubsection{Reward Function}
Policy optimization aims to optimize a model to maximize an arbitrary reward function.
Therefore, we design a \toward reward $T$ and an \away reward $A$ to mirror the \toward and \away objectives so that maximizing the rewards is equivalent to directly optimizing for the two objectives.
Specifically, we define $T$ and $A$ as
\begin{align*}
    T(\textbf{x}_{s\rightarrow t}, t)           &= SIM_{sty}(\textbf{x}_{s\rightarrow t}, t) \\
    A(\textbf{x}_{s\rightarrow t}, s)           &= 1 - SIM_{sty}(\textbf{x}_{s\rightarrow t}, s)
\end{align*}
where $SIM_{sty}$ is the style similarity calculated by the reward model $f_{reward}$.
However, our preliminary experiments show that training with only these two rewards sometimes results in models that only generate empty or very short outputs.
To mitigate this, we add the simple and quick-to-calculate length penalty term from \citet{wieting-etal-2019-beyond} to the reward, which is defined as
\begin{equation*}
    LP(\textbf{x}_{s\rightarrow t}, \textbf{x}_s) = e^{1-\frac{min(|\textbf{x}_{s\rightarrow t}|, |\textbf{x}_s|)}{max(|\textbf{x}_{s\rightarrow t}|, |\textbf{x}_s|)}}
\end{equation*}
The total reward is then
\begin{equation*}
    R = T + A - (LP^\alpha - 1)
\end{equation*}
where $\alpha$ is a temperature hyperparameter.

\subsubsection{PO Training Data}
During SFT, we train the transfer model to transfer the neutral paraphrase back into the original style before paraphrasing, which means the source style before paraphrasing and the target style are the same.
For PO, we want to further optimize the model to move the style of the transferred text away from the source style and toward the target style.
In this case, we have to make sure the source style and the target style are different, otherwise the two objectives will be contradictory to each other.
Therefore, during PO training, we shift the target style by one element, which yields a new dataset $\mathcal{T}_{PO} = \{(\textbf{y}_1, s_2), (\textbf{y}_2, s_3), \cdots, (\textbf{y}_n, s_1)\}$. Note that we also drop the gold outputs $\textbf{x}_i$ from $\mathcal{T}_{SFT}$ since PO trains the model on generated outputs and the rewards.

\subsubsection{PO Algorithms}
We consider three PO algorithms, Proximal Policy Optimization (PPO) \citep{Schulman2017ProximalPO}, Direct Preference Optimization (DPO) \citep{Rafailov2023DirectPO}, and Contrastive Preference Optimization (CPO) \citep{xu2024contrastive}.
PPO is an online reinforcement learning (RL) algorithm, while DPO and CPO are recent RL-free alternatives to PPO and have been shown to be more stable, computationally efficient, and effective on various NLP tasks \citep{Rafailov2023DirectPO, xu2024contrastive}.

\subsection{Inference}
For inference, given a text $\textbf{x}_s$ in style $s$, we transfer it into the target style $t$ by
\begin{equation*}
    \textbf{x}_{s\rightarrow t} = f_{PO}(f_{para}(\textbf{x}_s), t)
\end{equation*}
where $\textbf{x}_{s\rightarrow t}$ is the transferred text.

\section{Experiments}
We evaluate our approach on authorship style transfer tasks at two data resource levels, low-resource individual authorship style transfer and medium-resource community authorship style transfer.
In this section, we first discuss the task specific details for each task and then introduce the baseline models and implementation details.

\subsection{Individual Authorship Style Transfer}
In this section, we discuss the experiments on the individual authorship style transfer task.
Specifically, we adopt the same low-resource authorship style transfer task as in \citet{patel2023lowresource}, which aims to transfer a text from an arbitrary author into the style of another arbitrary author, for which only a limited number of text exemplars exist.

\subsubsection{Dataset}
We use the Million User Dataset (MUD) from \citet{khan-etal-2021-deep} to train and evaluate our model.
MUD is a dataset extracted from the Pushshift Reddit dataset \citep{DBLP:conf/icwsm/BaumgartnerZKSB20} which contains user posts on Reddit with author labels.
In this task, the author label is used as the style label $s$ in the dataset $\mathcal{D}$.
For training, we randomly sample 12,000 authors from the training split of MUD and use 10,000, 1,000, and 1,000 authors for training, validation, and test, respectively.
We randomly sample two texts for each author in the training split and one text for each author in the validation and the test splits.
For evaluation, we randomly sample 100 source authors and 100 target authors from the ``test\_query'' split of MUD; each author has 16 texts.

\subsubsection{Transfer Model Formulation}
In this task, we use a single model conditioned on few-shot target author exemplars for all authors, since \citet{patel2023lowresource} shows that in this extremely low-resource setting, exemplar-based approach works better than one model per author.

\subsubsection{Reward Model}
\label{Reddit_Reward_Model}
For policy optimization, we need a task specific reward model to calculate the style similarity score $SIM_{sty}(\textbf{x}, s)$ in the reward function.
For the individual-level transfer task, we use the LUAR model from \citet{rivera-soto-etal-2021-learning} as the reward model $f_{reward}$ which generates a single-vector authorship representation for an author with several texts from that author.
\begin{equation*}
    \textbf{v}_s = LU\!AR(\{\textbf{x}_i|s_i=s\})
\end{equation*}
We define the style similarity score as
\begin{equation*}
    SIM_{sty}(\textbf{x}, s) = cossim(LU\!AR(\{\textbf{x}\}), \textbf{v}_s))
\end{equation*}
where $cossim$ is the cosine similarity between the two vectors.

\subsubsection{Metrics}\label{sec:reddit_metrics}
We adopt a subset of automatic evaluation metrics from \citet{patel2023lowresource} to evaluate the style transfer, the content preservation, and the overall performance of our model.
For convenience, in this section, we denote the set of all test texts as $\mathcal{X}$ and the set of all test source-target author pairs as $\mathcal{S}$.

\noindent \textbf{Style Transfer}\quad We use the toward\footnote{For internal consistency, we refer to the `toward\textbf{s} score,' described in \citet{patel2023lowresource} as toward score in this work.}, away, and confusion scores to measure the style transfer performance.
The toward and away scores measure to what extent the authorship style transfer model moves the style of the texts away from the source style and toward the target style in the authorship representation space.
Concretely, the \textbf{toward} score is defined as
\begin{equation*}
    \frac{1}{|\mathcal{S}|} \sum_{(s, t)\in\mathcal{S}} \frac{1 - max(sim(\textbf{v}_{s\rightarrow t}, \textbf{v}_{s}), sim(\textbf{v}_{t}, \textbf{v}_{s}))}{1-sim(\textbf{v}_{t}, \textbf{v}_{s})}
\end{equation*}
and the \textbf{away} score is defined as
\begin{equation*}
    \frac{1}{|\mathcal{S}|} \sum_{(s, t)\in\mathcal{S}} \frac{max(sim(\textbf{v}_{s\rightarrow t}, \textbf{v}_{t}) - sim(\textbf{v}_{s}, \textbf{v}_{t}), 0)}{1-sim(\textbf{v}_{t}, \textbf{v}_{s})}
\end{equation*}
where $\textbf{v}_{s}$, $\textbf{v}_{t}$, and $\textbf{v}_{s\rightarrow t}$ are the LUAR authorship representations for the source author $s$, target author $t$, and the transferred texts, respectively, and $sim$ is a vector similarity measure from \citet{Cer2018UniversalSE}, which is defined as
\begin{equation*}
    sim(\textbf{u}, \textbf{v})=1 - \frac{arccos(\frac{\textbf{u}\cdot\textbf{v}}{\|\textbf{u}\|\|\textbf{v}\|})}{\pi}
\end{equation*}
We also use the confusion score to directly measure what percentage of the transferred texts is closer to the target style than the source style.
Formally, the \textbf{confusion} score is defined as
\begin{equation*}
    \frac{1}{|\mathcal{S}|} \sum_{(s, t)\in\mathcal{S}} \textbf{1}_{sim(\textbf{v}_{s\rightarrow t}, \textbf{v}_{t}) > sim(\textbf{v}_{s\rightarrow t}, \textbf{v}_{s})}
\end{equation*}

\noindent \textbf{Content Preservation}\quad To measure content preservation, we use the SBERT\footnote{We use the best-performing variant of SBERT, all-mpnet-base-v2.} \citep{reimers-gurevych-2019-sentence} cosine similarity instead of the mutual implication score (MIS) \citep{babakov-etal-2022-large} in \citet{patel2023lowresource} since SBERT is the most commonly used semantic embedding model, and MIS is trained on very short texts, but our test set contains much longer texts.
The \textbf{SBERT} content preservation score is simply
\begin{equation*}
    \frac{1}{|\mathcal{X}|} \sum_{\textbf{x}_s\in\mathcal{X}} cossim(\text{SBERT}(\textbf{x}_{s\rightarrow t}), \text{SBERT}(\textbf{x}_{s}))
\end{equation*}
where $\textbf{x}_{s}$ is the original text, and $\textbf{x}_{s\rightarrow t}$ is the transferred text.

\noindent \textbf{Overall Performance}\quad To have a better understanding of the overall performance of the models, we use the same method as in \citet{patel2023lowresource} to aggregate the toward, away, and the SBERT cosine similarity scores to obtain a \textbf{joint} score. Specifically,
\begin{equation*}
    joint = G(G(toward, away), cossim_{\text{SBERT}})
\end{equation*}
where $G$ refers to geometric mean.

\subsection{Community Authorship Style Transfer}
In the previous section, we investigated the effectiveness of our approach in transferring style across individual authors with an extremely limited number of texts. 
Different from such low-resource and fine-grained control, in this section, we demonstrate that our approach is equally proficient in transferring authorship style across communities of authors sharing the same attribute.
Specifically, we choose the native language ($L1$) of authors whose first language is not English as the attribute we want to control.
The objective is to take English text written by an author whose native language is $L1(s)$, say $s = Arabic$, and re-write it as English text in the style of a native $L1(t)$ author, say $t = Chinese$. 

\subsubsection{Dataset}
We use the ETS Corpus of Non-Native Written English\footnote{LDC2014T06} to study the $L1$ transfer task.
This ETS-TOEFL dataset has essays written by students whose $L1$ varies across 11 languages: Arabic, Chinese, French, German, Hindi, Italian, Japanese, Korean, Spanish, Telugu, and Turkish. 
The native language $L1$ is used as the style label $s$ in the dataset $\mathcal{D}$ in this task.
The data is carefully curated to control for topics: the topics are not correlated with $L1$ and all subjects write about the same set of topics. 
This removes the possibility for the system to make spurious topic-oriented correlations.\footnote{For example, if the topics are not controlled, the system could perhaps determine that the author’s likely $L1$ is either Hindi or Telugu if the topic centers around the game of Cricket.}
This is in contrast to other attribute specific datasets which typically do not control for unintended correlations between the categorical attributes and textual content.
The train, validation, and test splits have 900, 100, and 100 documents, respectively.

Our preliminary experiment shows that LLMs like GPT-3.5-turbo tend to drop information when paraphrasing long documents like the documents in the ETS dataset, so in this task, we process the documents at segment level.
Concretely, we split all documents into segments with up to 128 tokens, and the style labels of the segments are the same as the original documents.
We then sample 2,000 segments and 200 segments for each native language for training and validation, respectively, which results in a training set with 22,000 segments and a validation set with 2,200 segments.
For evaluation, we transfer all documents in the test set into all native language styles except the source style to obtain transferred texts for all 110 native language pairs.
This transfer is also done at segment level.
We segment all documents in the test set before transfer and regroup them back to documents afterward.

\subsubsection{Transfer Model Formulation}
In this task, we use a single model for each style since we have a fair amount of data for each style, and our preliminary experiments show that one model per author works better than the few-shot exemplar-based approach or control code-based \citep{Keskar2019CTRLAC} approach on this task.
We train the models using the STRAP approach, so the SFT model is the same as the STRAP model for this task.

\subsubsection{Reward Model}
\label{ETS_Reward_Model}
In the community-level transfer task, we use a classifier as the reward model $f_{reward}$ instead of the representation model since our preliminary experiment shows that the classifier works much better than the representation model on native language identification\footnote{Please see \autoref{sec:reward_model_selection} for detailed comparison.}.
Specifically, we train a RoBERTa-large \citep{liu2019roberta} classifier with 11 binary classification heads, corresponding to each native language on the ETS training set.
Formally, given an input text \textbf{x} and a topic ($L1$)  $s$, we denote the classifier output probability as $p_s(\textbf{x})$ and the classification decision as $C_s(\textbf{x})=\textbf{1}_{p_s(\textbf{x}) > 0.5}$.
We then define the style similarity score as
\begin{equation*}
    SIM_{sty}(\textbf{x}, s) = p_s(\textbf{x})
\end{equation*}

\subsubsection{Metrics}
We use the \textbf{SBERT} cosine similarity and the \textbf{joint} score defined in \autoref{sec:reddit_metrics} to evaluate the content preservation and the overall performance of our model.
However, since the representation model does not work well on the community authorship identification task, we propose three new metrics for style transfer accuracy in direct analogy with the $toward$, $away$, and $confusion$ scores in \citet{patel2023lowresource}.
For convenience, we denote all test documents written by authors with native language $s$ as $D_s$ in this section.

We use toward and away scores to indicate the percentage increase in how many transferred documents are classified as being written by a target native language author and the percentage decrease in how many transferred documents are classified as being written by a source native language author.
Formally, for each pair of source native language $s$ and target native language $t$, we define the \textbf{toward} score\footnote{The toward score and the \textit{\toward} objective/reward both measure to what extent the transferred texts reflect the target style, but they are defined differently. The toward score is defined to be more intuitive, while the \textit{\toward} objective/reward is defined to be easier to calculate. Similar for the away score.} as
\begin{equation*}
    max\left(\frac{\sum_{\textbf{x}_s\in D_s} C_t(\textbf{x}_{s\rightarrow t}) - \sum_{\textbf{x}_s\in D_s} C_t(\textbf{x}_s)}{|D_s| - \sum_{\textbf{x}_s\in D_s}C_t(\textbf{x}_s)}, 0\right)
\end{equation*}
and define the \textbf{away} score as
\begin{equation*}
    max\left(\frac{\sum_{\textbf{x}_s\in D_s} C_s(\textbf{x}_{s}) - \sum_{\textbf{x}_s\in D_{s}} C_s(\textbf{x}_{s\rightarrow t})}{\sum_{\textbf{x}_s\in D_s} C_s(\textbf{x}_{s})}, 0\right)
\end{equation*}
where $\textbf{x}_s$ referes to the original text and $\textbf{x}_{s\rightarrow t}$ refers to the transferred text.

We use the confusion score to measure what percentage of the transferred texts are classified as being written by a target native language author but not a source native language author.
Formally, the \textbf{confusion} score is defined as
\begin{equation*}
    \frac{\sum_{\textbf{x}_s\in D_s} \textbf{1}_{C_t(\textbf{x}_{s\rightarrow t}) - C_s(\textbf{x}_{s\rightarrow t}) = 1}}{|D_s|}
\end{equation*}

\subsection{Baseline Models}
We compare \name with a popular unsupervised style transfer model, STRAP \citep{krishna-etal-2020-reformulating}, the SOTA low-resource individual authorship style transfer model, STYLL \citep{patel2023lowresource}, the SOTA high-resource style transfer model STEER \citep{hallinan-etal-2023-steer}, and LLM zero-shot transfer.

\noindent\textbf{STRAP} performs text style transfer by paraphrasing the input text twice with a diverse paraphraser followed by an inverse paraphraser trained to rewrite the diverse paraphrase into the target style.

\noindent\textbf{STYLL} transfers the input text by prompting LLMs with the target style descriptors and few-shot transfer examples generated from the target style exemplars.

\noindent\textbf{STEER} trains the style transfer model with expert-guided data generation \citep{liu-etal-2021-dexperts} and a two-phase online-then-offline RL training using QUARK \citep{NEURIPS2022_b125999b}.

\noindent\textbf{Zero-shot Transfer} simply prompts LLMs with the input text and the target style to transfer.

\subsection{Implementation Details}
We implement our training framework and models with Huggingface's Transformers, PEFT, and TRL codebases.
Except GPT-3.5-turbo, LLaMA-2-7B-chat, and BLOOM-7B used for the zero-shot and STYLL baselines, we only use LLaMA-2-7B for all other approaches.
For computational efficiency, all learning-based models are trained with the Low-Rank Adaptation (LoRA) \citep{hu2022lora} technique.\footnote{Since STEER uses the QUARK algorithm which adds new tokens to the model, we also train the token embedding layer for STEER in the RL phase.}
Please see \autoref{sec:more_hyperparameters} and \autoref{sec:more_input_formats} for more details on the hyperparameters and the model input formats, respectively.

\begin{table*}[htb!]
\centering
\begin{tabular}{lccccc}
\toprule
Method & Toward        & Away           & SBERT          & \textbf{Joint}          & Confusion     \\
\midrule\midrule
STRAP  & 0.088$^\ddagger$          & {\ul 0.793}$^\dagger$    & 0.650$^\ddagger$          & 0.414$^\ddagger$          & 0.30$^\ddagger$         \\
STYLL  & 0.159          & \textbf{0.845} & 0.529$^\ddagger$          & 0.440$^\ddagger$          & \textbf{0.59} \\
SFT    & 0.137$^\ddagger$          & 0.707$^\ddagger$          & \textbf{0.754} & 0.484$^\ddagger$          & 0.32$^\ddagger$          \\
\midrule
\name-PPO    & 0.147$^\ddagger$          & 0.773$^\ddagger$          & 0.729$^\dagger$          & 0.495$^\dagger$          & {\ul 0.48}$^\dagger$    \\
\name-DPO    & {\ul 0.164}    & 0.748$^\ddagger$          & {\ul 0.733}$^\dagger$    & \textbf{0.507} & 0.44$^\dagger$          \\
\name-CPO    & \textbf{0.165} & 0.752$^\ddagger$          & 0.726$^\dagger$          & {\ul 0.505}$^\dagger$    & 0.46$^\dagger$          \\
\bottomrule
\end{tabular}
\caption{
    The automatic evaluation results on the individual authorship style transfer task with LLaMA-7B based models and \name models trained with the reward function $R = T + A - (LP^\alpha - 1)$.
    The best and the second best scores for each metric are shown in \textbf{bold} and {\ul underline}, respectively.
    "$\dagger$" and "$\ddagger$" indicate a significant difference between the model and the best model or the top two models, respectively, determined by t-test with $\alpha=0.05$.
}
\label{tab:reddit_agg}
\end{table*}

\section{Results}

In this section, we discuss and analyze the experimental results for both tasks.
Due to the limited time and computational resources, we conduct all experiments in a single run and perform statistical significance tests on the results.\footnote{Please see \autoref{sec:more_ttest} for details.}
For conciseness, we only show the automatic evaluation results with LLaMA-7B for all approaches and the results with the best reward combinations for \name.
Please see \autoref{sec:full_auto_eval} for the full automatic evaluation results and the ablation study.
We also conduct human evaluation and a case study for the community authorship transfer experiments.
Please see \autoref{sec:human_study} and \autoref{sec:case_study} for details.

\subsection{Individual Authorship Style Transfer}

The automatic evaluation results on the individual authorship style transfer task are shown in \autoref{tab:reddit_agg}.
We only show the \name models trained with the full reward function $R = T + A - (LP^\alpha - 1)$ since this reward function yields the best Joint score for all three PO algorithms on this task.
The joint score indicates that the overall performance of the three \name models are superior to all baseline models, and the two models trained with the RL-free PO algorithms (i.e. DPO and CPO) perform similarly to each other and both perform better than the RL-based PO algorithm (i.e. PPO).

Looking at the toward, away, and SBERT scores separately, we find that all PO algorithms can effectively improve the toward and the away scores, but at the cost of harming the SBERT score, since our reward function does not take semantic similarity into account, for efficiency and stability.
Even so, the three \name models still have decent SBERT scores that are higher than all baseline models except the SFT model since the KL-divergence penalty helps the models preserve the capability to keep the semantic meaning of the input texts.
One may notice that STYLL has much better away and confusion scores than all other models, but this is because the model sometimes copies some irrelevant content from the target exemplars which changes the meaning of the transferred texts, and this also explains why the SBERT score for STYLL is much lower than other models.

\subsection{Community Authorship Style Transfer}

The automatic evaluation results on the community authorship style transfer task are shown in \autoref{tab:ets_agg}.
For this task, the best \name models are trained with the reward function without the away reward $R = T - (LP^\alpha - 1)$.
The joint score indicates that DPO- and CPO-\name have the best overall performance.
They also have the best toward and confusion scores.
PPO can also slightly improve the performance of the SFT model, but the improvement is much less than DPO and CPO.
Similar to the previous task, PO training harms the SBERT score, but the magnitude of the loss is very small, and the result SBERT scores are still higher than all baseline models except STRAP/SFT.

\begin{table*}[htb!]
\centering
\begin{tabular}{lccccc}
\toprule
Method      & Toward        & Away           & SBERT          & \textbf{Joint}          & Confusion      \\
\midrule\midrule
Zero-shot   & 0.022$^\ddagger$          & 0.880$^\dagger$          & 0.738$^\ddagger$          & 0.321$^\ddagger$          & 0.033$^\ddagger$          \\
STYLL       & 0.210$^\ddagger$          & 0.832$^\ddagger$          & 0.854$^\ddagger$          & 0.598$^\ddagger$          & 0.227$^\ddagger$          \\
STRAP / SFT & 0.286$^\ddagger$          & 0.785$^\ddagger$          & \textbf{0.917} & 0.659$^\ddagger$          & 0.300$^\ddagger$          \\
STEER       & 0.334$^\ddagger$          & \textbf{0.926} & 0.879$^\ddagger$          & 0.699$^\ddagger$           & 0.348$^\ddagger$          \\
\midrule
\name-PPO         & 0.299$^\ddagger$          & 0.800$^\ddagger$          & 0.905$^\ddagger$          & 0.665$^\ddagger$          & 0.313$^\ddagger$          \\
\name-DPO         & {\ul 0.490}$^\dagger$    & 0.843$^\ddagger$          & {\ul 0.915}$^\dagger$    & {\ul 0.767}$^\dagger$    & {\ul 0.499}$^\dagger$    \\
\name-CPO         & \textbf{0.655} & {\ul 0.887}$^\dagger$    & 0.897$^\ddagger$          & \textbf{0.827} & \textbf{0.662} \\
\bottomrule
\end{tabular}
\caption{
    The automatic evaluation results on the community authorship style transfer task with LLaMA-7B based models and \name models trained with the reward function $R = T - (LP^\alpha - 1)$.
    The scores are averages over all pairs of native languages.
    The best and the second best scores for each metric are shown in \textbf{bold} and {\ul underline}, respectively.
    "$\dagger$" and "$\ddagger$" indicate a significant difference between the model and the best model or the top two models, respectively, determined by t-test with $\alpha=0.05$.
}
\label{tab:ets_agg}
\end{table*}

\section{Related Work}

\noindent\textbf{Text Style Transfer}\quad Since parallel data is very rare for text style transfer, only a few works solve this task in a supervised manner \citep{zhu-etal-2010-monolingual, rao-tetreault-2018-dear, kim-etal-2022-improving, raheja-etal-2023-coedit}.
Constrained by the datasets, these works only focus on some specific sub-tasks such as text simplification and formality transfer.
Therefore, to build more general style transfer models, recent works develop unsupervised methods that do not rely on parallel data.
These works mainly fall in five categories, content-style representation disentanglement \citep{Liu2019RevisionIC, jin-etal-2020-hooks}, style-related phrase replacement \citep{madaan-etal-2020-politeness, malmi-etal-2020-unsupervised, reid-zhong-2021-lewis}, reinforcement learning on direct objective \citep{gong-etal-2019-reinforcement, liu-etal-2021-learning, deng-etal-2022-rlprompt, hallinan-etal-2023-steer}, pseudo-parallel data generation \citep{krishna-etal-2020-reformulating, riley-etal-2021-textsettr}, and LLM prompting \citep{reif-etal-2022-recipe, suzgun-etal-2022-prompt, patel2023lowresource}.

The state-of-the-art authorship style transfer model, STYLL \citep{patel2023lowresource} transfers the input texts by prompting an LLM with the target style descriptors and few-shot pseudo-parallel transfer pairs generated by the same LLM, which combines the strength of pseudo-parallel generation and LLM prompting.
Even so, as a prompting-based method, STYLL can be potentially enhanced by RL since RL has already been shown to be effective in improving the performance of prompting-based style transfer models \citep{deng-etal-2022-rlprompt}, and the state-of-the-art general style transfer model, STEER \citep{hallinan-etal-2023-steer} is also trained with RL.
However, RL algorithms are shown to be unstable and hard to tune compared to the recently developed RL-free policy optimization algorithms such as DPO \citep{Rafailov2023DirectPO} and CPO \citep{xu2024contrastive}.

Therefore, in this work, we choose the solve the authorship style transfer task with a PO-based training framework.
Similar to STEER, we first generate pseudo-parallel data from the labeled non-parallel data and then train the model on the generated data, but our framework differs from STEER in three major ways: (1) we use a much simpler data generation strategy which only needs one paraphrase model and generates once for each instance in the non-parallel data, but STEER requires two extra models for each style as well as heavy overgeneration and filtering; (2) we only perform a single stage PO training instead of the two-stage offline-then-online RL training in STEER, and our reward function requires only one reward model instead of the three reward model in STEER; (3) we also use more stable and efficient RL-free PO algorithms instead of just the RL-based algorithm in STEER.

\noindent\textbf{Policy Optimization}\quad Policy optimization has been widely used in NLP to train language models on task specific objectives such as text simplification \citep{laban-etal-2021-keep}, question answering \citep{liu-etal-2022-rainier}, and machine translation \citep{xu2024contrastive}.
Most early works in this area focus on RL-based algorithms such as REINFORCE \citep{10.1007/BF00992696} and PPO \citep{Schulman2017ProximalPO}, but these algorithms are often considered unstable and inefficient.
Recently, many RL-free algorithms have been developed to improve the stability and the efficiency.
These works mostly focus on aligning LLMs with human preference \citep{Rafailov2023DirectPO, song2023preference}, but there are also some that apply to other tasks such as machine translation \citep{xu2024contrastive}.
In this work, we use PO algorithms to train the models directly on the authorship style transfer objectives.
To our best knowledge, this is the first work applying RL-free PO algorithms on text style transfer.

\section{Conclusion}

In this work, we propose a PO-based training framework for authorship style transfer, which combines the strength of supervised fine-tuning on the pseudo-parallel data and policy optimization on the transfer objective.
Extensive experiments confirm the effectiveness of our model on both low-resource and high-resource authorship style transfer tasks and show that our model outperforms the SOTA models in both authorship style transfer and general style transfer.
\section*{Limitations}

Although our approach shows strong performance on authorship style transfer, the performance on low-resource transfer is still much weaker than the performance on high-resource transfer.
There are two possible reasons.
First, we use small-scale datasets for both tasks due to the limited computational resources.
It is sufficient to model the coarse-grained community authorship styles but may be insufficient for the individual authorship styles.
Therefore, if more computational resources are available, future work can investigate whether more training data can help improve the performance of the low-resource authorship transfer models.
Second, our authorship information injection strategy may not be optimal.
We use a popular exemplar-based approach to inject the authorship information in the low-resource transfer task, but there may be more efficient approaches such as using continuous vectors instead of discrete tokens.
This is out of the scope of this work, but future work can explore more efficient information injection strategies for low-resource authorship style transfer.

Moreover, even though the two RL-free PO algorithms, DPO and CPO already show a much better performance than PPO, in this work, we only use them in an offline manner as in the original papers.
However, one can naturally enhance DPO and CPO with online data generated by the updated policy during training, which can potentially improve the performance of the models.
Therefore, future work can focus on improving the training framework with online DPO or CPO training.

\section*{Ethical Considerations}
Like other transfer learning LLMs, the quality of our model outputs highly depends on the quality of the underlying LLM and the training data.
In this work, we use the original LLaMA-2-7B model instead of the chat version to ensure the flexibility for training, but it also has a higher risk of generating toxic texts.
Also, the datasets we use contain unfiltered texts from the online forum Reddit and may also lead to unethical generation.
Therefore, for real-world applications, we suggest carefully filtering the training data and also using a post-generation filter to avoid outputting unethical texts.
As a PO-based training framework, one can also add some terms to the reward function to encourage the model to generate safe and ethical outputs.

Both datasets we use in this work contain texts with personal identifiable information (PII) and/or unethical words.
We do not remove profane texts and texts containing PIIs for human evaluation to maximally preserve the style and meaning of the texts.
Our human study protocol has been approved by an institutional review board.

Our model is intended for personal and authorized use such as building personal chatbots or authorship privatization, but we also recognize some potential harmful usage such as maliciously mimicking some individuals without authorization and intentionally generating texts in an offensive style.
Therefore, we suggest keeping all personal data locally to prevent malicious mimicking.
For text privatization, we suggest transferring to community-level authorship styles or styles mixed from multiple authors to prevent exposing the information of individual authors.
To maximally preclude any unintended use, we only permit the use of our approach on public datasets or with the explicit consent of the target authors.

\section*{Acknowledgements}
This research is supported in part by the Office of the Director of National Intelligence (ODNI), Intelligence Advanced Research Projects Activity (IARPA), via the HIATUS Program contract \#2022-22072200006, and in part by the Defense Advanced Research Projects Agency (DARPA) under Agreement No. HR00112490374. The views and conclusions contained herein are those of the authors and should not be interpreted as necessarily representing the official policies, either expressed or implied, of ODNI, IARPA, DARPA, or the U.S. Government. The U.S. Government is authorized to reproduce and distribute reprints for governmental purposes notwithstanding any copyright annotation therein.

\bibliography{anthology,custom}

\appendix

\section{More Experimental Results}
\label{sec:more_results}

\subsection{Full Automatic Evaluation Results}
\label{sec:full_auto_eval}
We show the full automatic evaluation results in \autoref{tab:reddit_agg_full} and \autoref{tab:ets_agg_full}.

\begin{table*}[htb!]
\small
\centering
\begin{tabular}{lllccccc}
\toprule
Method                 & Model                     & Reward                & Toward         & Away           & SBERT          & Joint          & Confusion     \\
\midrule\midrule
STRAP                  & LLaMA-7B                  & \multicolumn{1}{c}{-} & 0.088$^\ddagger$          & 0.793$^\ddagger$          & 0.650$^\ddagger$          & 0.414$^\ddagger$          & 0.30$^\ddagger$         \\
\midrule
\multirow{3}{*}{STYLL} & GPT-3.5-turbo             & \multicolumn{1}{c}{-} & 0.045$^\ddagger$          & {\ul 0.825}$^\dagger$    & 0.713$^\ddagger$          & 0.370$^\ddagger$          & 0.33$^\ddagger$          \\
                       & BLOOM-7B                  & \multicolumn{1}{c}{-} & 0.117$^\ddagger$          & 0.796$^\ddagger$          & 0.546$^\ddagger$          & 0.408$^\ddagger$          & 0.37$^\ddagger$          \\
                       & LLaMA-7B                  & \multicolumn{1}{c}{-} & 0.159          & \textbf{0.845} & 0.529$^\ddagger$          & 0.440$^\ddagger$          & \textbf{0.59} \\
\midrule
SFT                    & LLaMA-7B                  & \multicolumn{1}{c}{-} & 0.137$^\ddagger$          & 0.707$^\ddagger$          & 0.754$^\ddagger$          & 0.484$^\ddagger$          & 0.32$^\ddagger$          \\
\midrule\midrule
\multirow{3}{*}{\name-PPO}   & \multirow{3}{*}{LLaMA-7B} & $T + (LP^\alpha - 1)$                  & 0.119$^\ddagger$          & 0.753$^\ddagger$          & \textbf{0.767} & 0.480$^\ddagger$          & 0.29$^\ddagger$          \\
                       &                           & $A + (LP^\alpha - 1)$                  & 0.111$^\ddagger$          & 0.761$^\ddagger$          & 0.710$^\ddagger$          & 0.454$^\ddagger$          & 0.37$^\ddagger$         \\
                       &                           & $T + A + (LP^\alpha - 1)$                & 0.147$^\ddagger$          & 0.773$^\ddagger$          & 0.729$^\ddagger$          & 0.495$^\dagger$          & {\ul 0.48}$^\dagger$    \\
\midrule
\multirow{3}{*}{\name-DPO}   & \multirow{3}{*}{LLaMA-7B} & $T + (LP^\alpha - 1)$                  & 0.148$^\ddagger$          & 0.732$^\ddagger$          & {\ul 0.761}    & 0.500$^\dagger$          & 0.34$^\ddagger$          \\
                       &                           & $A + (LP^\alpha - 1)$                  & 0.135$^\ddagger$          & 0.739$^\ddagger$          & 0.729$^\ddagger$          & 0.479$^\ddagger$          & 0.35$^\ddagger$          \\
                       &                           & $T + A + (LP^\alpha - 1)$                & {\ul 0.164}    & 0.748$^\ddagger$          & 0.733$^\ddagger$          & \textbf{0.507} & 0.44$^\dagger$          \\
\midrule
\multirow{3}{*}{\name-CPO}   & \multirow{3}{*}{LLaMA-7B} & $T + (LP^\alpha - 1)$                  & 0.151$^\ddagger$          & 0.743$^\ddagger$          & 0.749$^\ddagger$          & 0.501$^\dagger$          & 0.38$^\ddagger$          \\
                       &                           & $A + (LP^\alpha - 1)$                  & 0.146$^\ddagger$          & 0.731$^\ddagger$          & 0.721$^\ddagger$          & 0.485$^\ddagger$          & 0.33$^\ddagger$          \\
                       &                           & $T + A + (LP^\alpha - 1)$                & \textbf{0.165} & 0.752$^\ddagger$          & 0.726$^\ddagger$          & {\ul 0.505}$^\dagger$    & 0.46$^\dagger$          \\
\bottomrule
\end{tabular}
\caption{
    The automatic evaluation results on the individual authorship style transfer task.
    The best and the second best scores for each metric are shown in \textbf{bold} and {\ul underline}, respectively.
    "$\dagger$" and "$\ddagger$" indicate that the model is significantly different from the best model or the top two models, respectively, determined by t-test with $\alpha=0.05$.
}
\label{tab:reddit_agg_full}
\end{table*}

\begin{table*}[htb!]
\small
\centering
\begin{tabular}{lllccccc}
\toprule
Method                     & Model                     & Reward                & Toward         & Away           & SBERT          & Joint          & Confusion      \\
\midrule\midrule
\multirow{2}{*}{Zero-shot} & GPT-3.5-turbo             & \multicolumn{1}{c}{-} & 0.005$^\ddagger$          & 0.811$^\ddagger$          & 0.885$^\ddagger$          & 0.240$^\ddagger$          & 0.013$^\ddagger$          \\
                           & LLaMA-7B                  & \multicolumn{1}{c}{-} & 0.022$^\ddagger$          & 0.880$^\dagger$          & 0.738$^\ddagger$          & 0.321$^\ddagger$         & 0.033$^\ddagger$          \\
\midrule
\multirow{2}{*}{STYLL}     & BLOOM-7B                  & \multicolumn{1}{c}{-} & 0.049$^\ddagger$          & 0.673$^\ddagger$          & 0.828$^\ddagger$          & 0.388$^\ddagger$          & 0.065$^\ddagger$          \\
                           & LLaMA-7B                  & \multicolumn{1}{c}{-} & 0.210$^\ddagger$          & 0.832$^\ddagger$          & 0.854$^\ddagger$          & 0.598$^\ddagger$          & 0.227$^\ddagger$          \\
\midrule
STRAP / SFT               & LLaMA-7B                  & \multicolumn{1}{c}{-} & 0.286$^\ddagger$          & 0.785$^\ddagger$          & \textbf{0.917} & 0.659$^\ddagger$          & 0.300$^\ddagger$          \\
\midrule
STEER                      & LLaMA-7B             & $TSS + F + MS$\footnotemark & 0.334$^\ddagger$          & \textbf{0.926} & 0.879$^\ddagger$          & 0.699$^\ddagger$           & 0.348$^\ddagger$          \\
\midrule\midrule
\multirow{3}{*}{\name-PPO}       & \multirow{3}{*}{LLaMA-7B} & $T + (LP^\alpha - 1)$                & 0.299$^\ddagger$          & 0.800$^\ddagger$          & 0.905$^\ddagger$          & 0.665$^\ddagger$          & 0.313$^\ddagger$          \\
                           &                           & $A + (LP^\alpha - 1)$                & 0.235$^\ddagger$          & 0.782$^\ddagger$          & 0.906$^\ddagger$          & 0.623$^\ddagger$          & 0.250$^\ddagger$          \\
                           &                           & $T + A + (LP^\alpha - 1)$            & 0.240$^\ddagger$          & 0.788$^\ddagger$          & 0.908$^\ddagger$          & 0.628$^\ddagger$          & 0.256$^\ddagger$          \\
\midrule
\multirow{3}{*}{\name-DPO}       & \multirow{3}{*}{LLaMA-7B} & $T + (LP^\alpha - 1)$                & 0.490$^\ddagger$          & 0.843$^\ddagger$          & 0.915$^\ddagger$          & 0.767$^\ddagger$          & 0.499$^\ddagger$          \\
                           &                           & $A + (LP^\alpha - 1)$                & 0.321$^\ddagger$          & 0.789$^\ddagger$          & {\ul 0.917}    & 0.679$^\ddagger$          & 0.334$^\ddagger$          \\
                           &                           & $T + A + (LP^\alpha - 1)$            & 0.488$^\ddagger$          & 0.837$^\ddagger$          & 0.915$^\ddagger$          & 0.765$^\ddagger$          & 0.497$^\ddagger$          \\
\midrule
\multirow{3}{*}{\name-CPO}       & \multirow{3}{*}{LLaMA-7B} & $T + (LP^\alpha - 1)$                & \textbf{0.655} & 0.887$^\ddagger$          & 0.897$^\ddagger$          & \textbf{0.827} & \textbf{0.662} \\
                           &                           & $A + (LP^\alpha - 1)$                & 0.456$^\ddagger$          & 0.835$^\ddagger$          & 0.909$^\ddagger$          & 0.749$^\ddagger$          & 0.467$^\ddagger$          \\
                           &                           & $T + A + (LP^\alpha - 1)$            & {\ul 0.654}    & {\ul 0.891}$^\dagger$    & 0.896$^\ddagger$          & {\ul 0.827}    & {\ul 0.660}   \\
\bottomrule
\end{tabular}
\caption{
    The automatic evaluation results on the community authorship style transfer task.
    The scores are averages over all pairs of native languages.
    The best and the second best scores for each metric are shown in \textbf{bold} and {\ul underline}, respectively.
    "$\dagger$" and "$\ddagger$" indicate that the model is significantly different from the best model or the top two models, respectively, determined by t-test with $\alpha=0.05$.
}
\label{tab:ets_agg_full}
\end{table*}
\footnotetext{TSS, F, and MS are the three components of the reward function for STEER, which stand for target style strength, fluency, and meaning similarity, respectively.}

\subsection{More Baseline LLMs}
In addition to LLaMA-7B, we evaluate STYLL on BLOOM-7B since it has the best joint in \citep{patel2023lowresource}.
We also evaluate STYLL on GPT-3.5-turbo since the GPT-3 endpoint used in \citep{patel2023lowresource} is deprecated by OpenAI, and GPT-3.5-turbo is the closest available model, but we only use it for the individual authorship transfer task due to the limited budget.
For the zero-shot transfer approach, we also use GPT-3.5-turbo to show its performance with one of the current best LLMs.
We do not use GPT-4 due to the limited budget.
Compared to BLOOM-7B and GPT-3.5-turbo, LLaMA-7B has the best joint score in all baseline approaches on both tasks, so the full results are still consistent with the concise version in \autoref{tab:reddit_agg} and \autoref{tab:ets_agg}.

\subsection{Reward Function Ablation Study}
We ablate the toward reward and the away reward from the reward function separately to assess their individual effects on the model performance.
For the individual authorship transfer task, when using partial reward functions without the toward reward or the away reward, the PO algorithms can still improve the score corresponding to the remaining term in the reward function in most cases, but none of the towards, away, and joint scores is as good as the model trained on the full reward function using each algorithm.
In contrast, for the community authorship transfer task, the away reward does not help improve the away score in most cases, and training with only the toward reward and the length penalty yields the model with the best overall performance for each PO algorithm.

\subsection{Human Evaluation}
\label{sec:human_study}

We conduct a human study on the community authorship transfer task.
We randomly select 10 samples for each target native language from the test set for STYLL, STRAP, STEER, and \name-CPO, and collect up to 3 annotations for each.
The samples are evaluated in two dimensions, style confusion (SC) and content preservation (CP).
The style confusion is a simpler version of the confusion score we use in the automatic evaluation.
We show the annotators three examples each in the source style and the target style, and ask them ``which is the style of the transferred text''.
The confusion score is 1 if they select the target style, other with 0.
We assess the content preservation using a 3-point Likert scale ranging from 0 to 2.
The detailed instructions are shown in \autoref{tab:human_eval_inst}.

The results are shown in \autoref{tab:human_eval}.
Since the style classification task has been shown to be very difficult for humans \cite{krishna-etal-2020-reformulating, hallinan-etal-2023-steer}, we perform an independent t-test on the results and find no statistically significant difference in style confusion in any model pairs.
However, we observe statistically significant differences in content preservation, which indicates that both STRAP and STEER are significantly better than STYLL and \name-CPO.
Also, the results on content preservation are generally consistent with the SBERT in the automatic evaluation except for the STEER model.

\begin{table}[]
\centering
\begin{tabular}{l@{\hspace{30pt}}cc}
\toprule
             & SC & CP \\
\midrule
STYLL        & {\ul 0.622}     & 0.955$^\ddagger$                \\
STRAP        & 0.516           & {\ul 1.267}          \\
STEER        & \textbf{0.690}  & \textbf{1.279}       \\
\name-CPO & 0.537           & 1.018$^\ddagger$               \\
\bottomrule
\end{tabular}
\caption{
    Human evaluation results on the community authorship transfer task.
    The best and the 2nd best scores in each column are emphasized in \textbf{bold} and {\ul underline}, respectively.
    ``$\ddagger$'' indicates a statistically significant difference between the top two models determined by independent sample t-test with $p<0.05$.
}
\label{tab:human_eval}
\end{table}

\begin{table*}[]
\centering
\begin{tabularx}{\textwidth}{llX}
\toprule
Style Confusion                       & Question         & Based on the examples above, what is the style of the following text?                                                                     \\
\midrule
\multirow{9}{*}{Content Preservation} & Similar          & Most of the meaning (75\% or more) of the two passages is the same.                                                                       \\
\cmidrule(rl){2-3}
                                      & Somewhat Similar & Large portions (50-75\%) of the passages are the same, but there are significant sections that differ or are present in only one passage. \\
\cmidrule(rl){2-3}
                                      & Not Similar      & Only small portions (less than 50\%) of the passages are the same.                                                                        \\
\cmidrule(rl){2-3}
                                      & Question         & How similar are the following two texts?     \\
\bottomrule
\end{tabularx}
\caption{
    Instructions used in the human evaluation.
}
\label{tab:human_eval_inst}
\end{table*}

\subsection{Case Study}
\label{sec:case_study}

\begin{table*}[htb!]
\small
\centering
\begin{tabularx}{\textwidth}{lX}
\toprule
Target  Sample                & traviling is a very nice thing , it hepls you see new coultcurs and to meet new poeplo , there are alot of ways for traviling  and i believe that the best way is to travil with a group led by a tour guide , [... more] \\ 
\midrule\midrule
Source Text                   & The influnce of advertisements on the customers is worth commendable.  The advertisers are projecting thier goods to the customers in a \textquotesingle larger than real{\textquotesingle} manner.{\textquotesingle} [... more]                                           \\
\midrule
Zero-shot                     & Advertisements have a powerful influence on consumers, overstating the features and benefits of products to make them seem better than they actually are. [... more]                                                     \\
\midrule
STYLL                         & Advertisements are used to promote products, making them appear attractive and useful. Through exaggeration, advertisers present their products as having exceptional features. [... more]                               \\
\midrule
STRAP                         & The advertisements have a great effect on the customers and they should be praised. The advertisers make the products seem more better in the eyes of the customers rather than they really are. [... more]              \\
\midrule
STEER                         & Advertising has a positive impact on customers, as it promotes products in a way that exaggerates their qualities. Advertisers often portray their products as superior to reality. [... more]                           \\
\midrule
\name-CPO                  & because the advertisment have  alot effec on the custumers , and make the products seem better than how they really are, [... more] \\
\bottomrule
\end{tabularx}
\caption{
    An example from the ETS test set. Due to the limited space, we only show the beginning of each document.
}
\label{tab:ets_ex}
\end{table*}

We show a transfer example on the community authorship transfer task in \autoref{tab:ets_ex} for a simple qualitative case study.
It shows that \name-CPO successfully captures a common typo, ``alot'' and three main characteristics of the target style: using all lowercase, using space before comma, and high typo rate, while no other model is able to capture any of these.
\section{More Implementation Details}
\label{sec:more_implementation_details}

\subsection{Statistical Significance Test}
\label{sec:more_ttest}

We perform a resampled paired t-test on all results.
Specifically, we randomly draw subsets from the test set and perform paired t-tests on the scores of the subsets.
For the individual authorship style transfer task, we sample at the author level since the style model works at the author level.
For the community authorship style transfer task, we sample at the document level.
The hyperparameters for the resampling t-test are shown in \autoref{tab:ttest}.

\begin{table}[]
\centering
\begin{tabular}{lcr@{\hspace{1mm}}l}
\toprule
           & \# Subsets & \multicolumn{2}{r}{Subset Size} \\
\midrule
Individual & 10         & 20            & authors         \\
Community  & 10         & 1100          & docs            \\
\bottomrule
\end{tabular}
\caption{
    Hyperparameters for the resampling t-test.
}
\label{tab:ttest}
\end{table}

\subsection{Hyperparameters}

\label{sec:more_hyperparameters}

\begin{table}[!ht]
\centering
\begin{tabular}{ll}
\toprule
LoRA Hyperparameters \\
\midrule
r              & 16               \\
$\alpha$          & 32               \\
dropout        & 0.05             \\
target modules & q\_proj, v\_proj \\
\bottomrule
\end{tabular}
\caption{
    Hyperparameters for the LoRA adapters.
}
\label{tab:lora_params}
\end{table}

\begin{table*}[!ht]
\centering
\begin{tabular}{lcccccc}
\toprule
               & Paraphraser & \multicolumn{4}{c}{\name} & STRAP \\
\cmidrule(rl){2-2}\cmidrule(rl){3-6}\cmidrule(rl){7-7}
               & SFT         & SFT    & PPO       & DPO   & CPO   & SFT   \\
\midrule\midrule
learning rate  & 5e-5        & 5e-5   & 1.41e-5   & 2e-6  & 2e-6  & 5e-5  \\
batch size     & 32          & 32     & 32        & 32    & 32    & 5     \\
\# epochs      & 6           & 6      & 6         & 6     & 6     & 60    \\
KL coef / $\beta$ & -           & -      & 0.2       & 0.5   & 0.1   & -     \\
top p          & -           & -      & 1.0       & 1.0   & 1.0   & -     \\
temperature    & -           & -      & 1.0       & 1.0   & 1.0   & -     \\
length penalty $\alpha$    & -           & -      & 0.5       & 0.5   & 0.5   & -     \\
\bottomrule
\end{tabular}
\caption{
    Training hyperparameters for the individual authorship style transfer task.
}
\label{tab:reddit_params}
\end{table*}

\begin{table*}[!ht]
\centering
\scriptsize
\begin{tabular}{lcccccccc}
\toprule
               & Paraphraser / Classifier & \multicolumn{4}{c}{\name} & STRAP & \multicolumn{2}{c}{STEER}                \\
\cmidrule(rl){2-2}\cmidrule(rl){3-6}\cmidrule(rl){7-7}\cmidrule(rl){8-9}
               & SFT         & SFT    & PPO      & DPO   & CPO   & SFT   & Expert Model & QUARK (RL)                \\
\midrule\midrule
learning rate  & 5e-5        & 5e-5   & 1.41e-5  & 2e-6  & 2e-6  & 5e-5  & 5e-5         & 5e-5                      \\
batch size     & 32          & 8      & 32       & 16    & 16    & 8     & 8            & 8                         \\
\# epochs      & 6           & 6      & 6        & 10    & 10    & 6     & 6            & 6 (offline) + 10 (online) \\
KL coef / $\beta$ & -           & -      & 0.2      & 0.5   & 0.1   & -     & -            & 0.025                     \\
top p          & -           & -      & 1.0      & 1.0   & 1.0   & -     & -            & 1.0                       \\
temperature    & -           & -      & 1.0      & 1.0   & 1.0   & -     & -            & 1.0                       \\
length penalty $\alpha$    & -           & -      & 0.5      & 0.5   & 0.5   & -     & -            & -                         \\
\bottomrule
\end{tabular}
\caption{
    Training hyperparameters for the community authorship style transfer task.
}
\label{tab:ets_params}
\end{table*}

\begin{table*}[!ht]
\centering
\begin{tabular}{lccc}
\toprule
                & \multicolumn{2}{c}{Pseudo-Parallel Data Generation}     & Inference  \\
\cmidrule(rl){2-3}\cmidrule(rl){4-4}
                & \name & STEER                   & All Models \\
\midrule\midrule
top p           & 1.0           & 1.0                     & 1.0        \\
temperature     & 0.7           & 1.0                     & 0.7        \\
DExperts $\alpha$           & -             & 1.0, 1.6, 1.8, 2.0, 2.2 & -          \\
over generation & $\times1$            & $\times50$                         & -      \\
\bottomrule
\end{tabular}
\caption{
    Generation hyperparameters for both tasks.
}
\label{tab:gen_params}
\end{table*}

\begin{table*}[!ht]
\centering
\begin{tabular}{llll}
\toprule
    & Learning Rate                      & KL coef / $\beta$     & Batch Size \\
\midrule\midrule
PPO & 1.41e-5, 2.82e-5, 4.23e-5          & 0.2                & 8, 16, 32  \\
DPO & 5e-7, 1e-6, 2e-6                   & 0.1, 0.2, 0.3, 0.5 & 8, 16, 32  \\
CPO & 5e-7, 1e-6, 2e-6, 5e-6, 1e-5, 1e-4 & 0.1, 0.5           & 8, 16, 32  \\
\bottomrule
\end{tabular}
\caption{
    Hyperparameters tested for \name.
}
\label{tab:po_hyperparam_search}
\end{table*}

\begin{table*}[!ht]
\centering
\begin{tabular}{lll}
\toprule
Phase                                               & Hyperparameter       &                                        \\
\midrule
\multirow{3}{*}{Expert-Guided Data Generation} & $\alpha$                 & 0.2, 0.4, 0.6, 1.0, 1.6, 1.8, 2.0, 2.2 \\
                                               & temperature           & 0.7, 1.0, 1.3                          \\
                                               & over generation        & $\times10$, $\times30$, $\times50$                            \\
\midrule
\multirow{3}{*}{QUARK (RL)}                         & Learning Rate         & 1e-5, 5e-5                             \\
                                               & KL coef               & 0.025, 0.05                            \\
                                               & Batch Size            & 8, 32                                  \\
\bottomrule
\end{tabular}
\caption{
    Hyperparameters tested for STEER.
}
\label{tab:steer_hyperparam_search}
\end{table*}

Due to  limited time and computational resources, we are not able to perform a thorough search on all hyperparameters, but we search for several important hyperparameters and show the best-performing hyperparameters for both training and generation in \autoref{tab:lora_params}, \autoref{tab:reddit_params}, \autoref{tab:ets_params}, and \autoref{tab:gen_params}.
\autoref{tab:po_hyperparam_search} and \autoref{tab:steer_hyperparam_search} show the hyperparameter we test in the experiments.
For the low-resource STRAP and all few-shot exemplar-based models, we use 5 target exemplars for each author.

\subsection{Model Input Formats}
\label{sec:more_input_formats}

\begin{table*}[]
\centering
\begin{tabularx}{\textwidth}{llX}
\toprule
Method                                                                         &                        Model                                                       & Prompt                                                                                                                                                                                                                                                                                                                                                                                                                                                                                                                                                                                                                                                                                                                                                  \\
\midrule\midrule
\multirow{1}{*}{Zero-shot}                                                    & LLaMA-7B                                                                      & {[}INST{]} \textless{}\textless{}SYS\textgreater{}\textgreater{}\textbackslash{}nYou are a college student whose native language is \textless{}target\_native\_language\textgreater{}.\textbackslash{}n\textless{}\textless{}/SYS\textgreater{}\textgreater{}\textbackslash{}n\textbackslash{}nUsing the writing style of a college student whose native language is \textless{}target\_native\_language\textgreater accurately paraphrase the following passage in English.\textbackslash{}n\textbackslash{}nOriginal Passage:\textbackslash{}n\textless{}text\_to\_be\_transferred\textgreater {[}/INST{]}Sure, using the writing style of a native \textless{}text\textgreater speaker, here is the paraphrased passage in English:\textbackslash{}n \\
\cmidrule{2-3}
                                                                              & GPT-3.5-turbo                                                                 & Passage: \textless{}text\textgreater{}\textbackslash{}n\textbackslash{}nUsing the writing style of a college student whose native language is \textless{}target\_native\_language\textgreater accurately paraphrase the passage in English.\textbackslash{}n\textbackslash{}nRewrite:                                                                                                                                                                                                                                                                                                                                                                                                                                                                   \\
\midrule
\multirow{2}{*}{\begin{tabular}[c]{@{}l@{}}STYLL\\ (paraphrase)\end{tabular}} & LLaMA-7B                                                                      & {[}INST{]} \textless{}\textless{}SYS\textgreater{}\textgreater{}\textbackslash{}nYou are an expert at paraphrasing.\textbackslash{}n\textless{}\textless{}/SYS\textgreater{}\textgreater{}\textbackslash{}n\textbackslash{}nPlease paraphrase the following passage in a simple neutral style.\textbackslash{}n\textbackslash{}n Passage: \textless{}text\textgreater {[}/INST{]}Sure! Here's a paraphrased version of the passage in a simple and neutral style:\textbackslash{}n\textbackslash{}n                                                                                                                                                                                                                                                     \\
\cmidrule{2-3}
                                                                              & GPT-3.5-turbo                                                                 & Passage: \textless{}text\textgreater{}\textbackslash{}n\textbackslash{}nParaphrase the passage in a simple neutral style.\textbackslash{}n\textbackslash{}nRewrite:                                                                                                                                                                                                                                                                                                                                                                                                                                                                                                                                                                                     \\
\midrule
\multirow{2}{*}{\begin{tabular}[c]{@{}l@{}}STYLL\\ (descriptor)\end{tabular}} & LLaMA-7B                                                                      & {[}INST{]} \textless{}\textless{}SYS\textgreater{}\textgreater{}\textbackslash{}nYou are an expert at writing style analysis.\textbackslash{}n\textless{}\textless{}/SYS\textgreater{}\textgreater{}\textbackslash{}n\textbackslash{}nPassage: \textless{}target\_text1\textgreater{}\textbackslash{}nPassage: \textless{}target\_text2\textgreater{}\textbackslash{}nList 5 adjectives, comma-separated, that describe the writing style of the author of these passages. {[}/INST{]}Sure, here are 5 adjectives, comma-separated, that describe the writing style of the author of these passages:                                                                                                                                                    \\
\cmidrule{2-3}
                                                                              & \begin{tabular}[t]{@{}l@{}}GPT-3.5-turbo \\ BLOOM-7B\end{tabular}             & Passage: \textless{}target\_text1\textgreater{}\textbackslash{}nPassage: \textless{}target\_text2\textgreater{}\textbackslash{}nList some adjectives, comma-separated, that describe the writing style of the author of these passages:                                                                                                                                                                                                                                                                                                                                                                                                                                                                                                                 \\
\midrule
\begin{tabular}[t]{@{}l@{}}STYLL (transfer)\end{tabular}                    & \begin{tabular}[t]{@{}l@{}}LLaMA-7B \\ GPT-3.5-turbo \\ BLOOM-7B\end{tabular} & Here is some text: \{\textless{}neutral\_target\_text1\textgreater{}\} Here is a rewrite of the text that is more \textless{}descriptors\textgreater{}: \{\textless{}target\_text1\textgreater{}\} Here is some text: \{\textless{}neutral\_target\_text2\textgreater{}\} Here is a rewrite of the text that is more \textless{}descriptors\textgreater{}: \{\textless{}target\_text2\textgreater{}\} Here is some text: \{\textless{}neutral\_source\_text\textgreater{}\} Here is a rewrite of the text that is more \textless{}descriptors\textgreater{}: \{  
\\
\bottomrule
\end{tabularx}
\caption{
    Prompts for the LLM prompting approaches. For exemplar-based approaches, we only show two target exemplars for illustration. Please see \autoref{sec:more_hyperparameters} for the actual number used in the experiments.
}
\label{tab:llm_prompts}
\end{table*}

\begin{table*}[]
\centering
\begin{tabular}{lll}
\toprule
Approach                                                                      & Level                                                          & Prompt                                                                                                                                                                                                                         \\
\midrule\midrule
\begin{tabular}[t]{@{}l@{}}Ours\\ (paraphrase)\end{tabular}                & \begin{tabular}[t]{@{}l@{}}Individual\\ Community\end{tabular} & {[}SRC{]}\textless{}text\textgreater{}{[}/SRC{]}                                                                                                                                                                               \\
\midrule
\multirow{2}{*}{\begin{tabular}[c]{@{}l@{}}Ours\\ (transfer)\end{tabular}} & Individual                                                     & \begin{tabular}[t]{@{}l@{}}{[}REF{]}\textless{}target\_text1\textgreater{}{[}/REF{]}{[}REF{]}\textless{}target\_text2\textgreater{}{[}/REF{]}\\ 
{[}SRC{]}\textless{}neutral\_source\_text\textgreater{}{[}/SRC{]}\end{tabular} \\
\cmidrule{2-3}
                                                                           & Community                                                      & {[}SRC{]}\textless{}neutral\_source\_text\textgreater{}{[}/SRC{]}                                                                                                                                                              \\
\midrule
\begin{tabular}[t]{@{}l@{}}STRAP\\ (paraphrase)\end{tabular}               & \begin{tabular}[t]{@{}l@{}}Individual\\ Community\end{tabular} & {[}SRC{]}\textless{}text\textgreater{}{[}/SRC{]}                                                                                                                                                                               \\
\midrule
\begin{tabular}[t]{@{}l@{}}STRAP\\ (transfer)\end{tabular}                 & \begin{tabular}[t]{@{}l@{}}Individual\\ Community\end{tabular} & {[}SRC{]}\textless{}neutral\_source\_text\textgreater{}{[}/SRC{]}                                                                                                                                                              \\
\midrule
STEER                                                                      & Community                                                      & {[}SRC{]}\textless{}source\_text\textgreater{}{[}/SRC{]}                                                                                                         \\
\bottomrule
\end{tabular}
\caption{
    Prompts for the learning-based approaches. For exemplar-based approaches, we only show two target exemplars for illustration. Please see \autoref{sec:more_hyperparameters} for the actual number used in the experiments.
}
\label{tab:learning_prompts}
\end{table*}

For LLM prompting based approaches, we use natural language prompts shown in \autoref{tab:llm_prompts}.
For learning-based approaches, we use simpler prompts with special tokens\footnote{We do not add new tokens to the tokenizer and model. All inputs are tokenized by the original tokenizer.} which are shown in \autoref{tab:learning_prompts}.

\subsection{Hardware and Runtime}
We report the training hardware and runtime for all learning-based approaches in this work in \autoref{tab:hardware_runtime}.

\begin{table*}[]
\centering
\begin{tabular}{llcccc}
\toprule
                       &              & \multicolumn{2}{c}{Individual} & \multicolumn{2}{c}{Community} \\
\cmidrule(rl){3-4}\cmidrule(rl){5-6}
                       &              & GPUs        & Time (hrs)       & GPUs        & Time (hrs)      \\
\midrule\midrule
Paraphraser            & SFT          & A40x2       & 2                & A40x2       & 3               \\
\midrule
\multirow{4}{*}{\name}          & SFT          & A40x2       & 12               & A40x1       & 6               \\
                       & PPO          & A40x2       & 29               & A40x2       & 40              \\
                       & DPO          & A40x2       & 4 + 8            & A40x2       & 9 + 14              \\
                       & CPO          & A40x2       & 4 + 6            & A40x2       & 9 + 10              \\
\midrule                       
STRAP                  & SFT          & A40x1       & 1                & A40x1       & 6               \\
\midrule
\multirow{2}{*}{STEER} & Expert Model & -           & -                & A40x1       & 3               \\
                       & QUARK (RL)   & -           & -                & A40x1       & 651 + 43              \\
\bottomrule
\end{tabular}
\caption{
    Training hardware and runtime for the learning-based approaches. For DPO, CPO, and STEER QUARK, the two numbers for time are the data generation time on a single A40 GPU and the training time, respectively. We do not report the runtime for the paraphrase data generation since it is done through the OpenAI API.
}
\label{tab:hardware_runtime}
\end{table*}

\subsection{Paraphraser Selection}
\label{sec:paraphraser_selection}
We train a LLaMA paraphraser on GPT-3.5-turbo generated paraphraser data for two main reasons.
First, the transfer pipeline is fully local with the LLaMA paraphraser, which is more cost-efficient and manageable.
Second, more importantly, we find that the LLaMA paraphraser trained on the GPT-3.5-turbo generated data performs even better than GPT-3.5-turbo.
Specifically, the trained LLaMA paraphraser achieves 0.764 SBERT cosine similarity on the test set, while GPT-3.5-turbo only has 0.738.

\subsection{Reward Model Selection}
\label{sec:reward_model_selection}

We use LUAR for the experiments on the Reddit data since LUAR is the SOTA and most widely used authorship verification model on the Reddit dataset.
However, preliminary experiments show that LUAR has only 0.53 accuracy on the ETS test set, while a trained classifier achieves 0.71 accuracy on the same test set.
Therefore, we use the trained classifier as the reward model for the experiments on the ETS dataset.
The reason for the large performance gap is that ETS has a countable number of classes (11 native languages) with plenty of training data to learn to accurately determine the author's native language.
Also, LUAR is a representation model that is designed to solve open-set problems in which the test data may have authors with textual styles never seen in the training collection, and this is in line with the setting of our low-resource transfer task on Reddit.

\section{Scientific Artifacts}
\label{sec:artifacts}

\subsection{Use of Existing Artifacts}

We list all existing artifacts we use in this work with their licenses and links in \autoref{tab:artifacts}.
The numbers of parameters of the models are shown in the same table in parentheses.
The artifacts are under various licenses, but all permit the use for research purposes.
All artifacts listed are allowed to be used in this work.

\begin{table*}[!ht]
\centering
\footnotesize
\begin{tabular}{lllL}
\toprule
Type                      & Name                                     & License           & \footnotesize Link                                                 \\
\midrule\midrule
\multirow{2}{*}{Dataset}  & Million User Dataset                     & Apache-2.0        & \href{https://github.com/noa/naacl2021}{https://github.com/noa/naacl2021}                                   \\
                          & ETS Corpus                               & \href{https://catalog.ldc.upenn.edu/license/ets-corpus-of-non-native-written-english.pdf}{License link}              & \href{https://catalog.ldc.upenn.edu/LDC2014T06}{https://catalog.ldc.upenn.edu/LDC2014T06}                           \\
\midrule
\multirow{8}{*}{Model}    & LLaMA-2-7B (6.7B)                              & Meta              & \href{https://huggingface.co/meta-llama/Llama-2-7b-hf}{https://huggingface.co/meta-llama/Llama-2-7b-hf}                    \\
                          & LLaMA-2-7B-chat (6.7B)                         & Meta              & \href{https://huggingface.co/meta-llama/Llama-2-7b-chat-hf}{https://huggingface.co/meta-llama/Llama-2-7b-chat-hf}              \\
                          & BLOOM-7B (7.1B)                                & RAIL License v1.0 & \href{https://huggingface.co/bigscience/bloom-7b1}{https://huggingface.co/bigscience/bloom-7b1}                        \\
                          & GPT-3.5-turbo (-)                           & MIT               & \href{https://platform.openai.com/docs/models/gpt-3-5-turbo}{https://platform.openai.com/docs/models/gpt-3-5-turbo}              \\
                          & RoBERTa-large (355M)                           & MIT               & \href{https://huggingface.co/FacebookAI/roberta-large}{https://huggingface.co/FacebookAI/roberta-large}                    \\
                          & RoBERTa-large-COLA (355M)                      & MIT               & \href{https://huggingface.co/cointegrated/roberta-large-cola-krishna2020}{https://huggingface.co/cointegrated/roberta-large-cola-krishna2020} \\
                          & all-mpnet-base-v2 (109M)                       & Apache-2.0        & \href{https://huggingface.co/sentence-transformers/all-mpnet-base-v2}{https://huggingface.co/sentence-transformers/all-mpnet-base-v2}     \\
                          & LUAR-MUD (83M)                                & Apache-2.0        & \href{https://huggingface.co/rrivera1849/LUAR-MUD}{https://huggingface.co/rrivera1849/LUAR-MUD}                        \\
\midrule
\multirow{6}{*}{Software} & Huggingface Transformers                 & Apache-2.0        & \href{https://github.com/huggingface/transformers}{https://github.com/huggingface/transformers}                        \\
                          & Huggingface PEFT                         & Apache-2.0        & \href{https://github.com/huggingface/peft}{https://github.com/huggingface/peft}                                \\
                          & Huggingface TRL                          & Apache-2.0        & \href{https://github.com/huggingface/trl}{https://github.com/huggingface/trl}                                 \\
                          & Sentence Transformers                    & Apache-2.0        & \href{https://github.com/UKPLab/sentence-transformers}{https://github.com/UKPLab/sentence-transformers}                    \\
                          & NLTK                                     & Apache-2.0        & \href{https://github.com/nltk/nltk}{https://github.com/nltk/nltk} \\
                          & ALMA (for CPO trainer)                                    & MIT        & \href{https://github.com/fe1ixxu/ALMA}{https://github.com/fe1ixxu/ALMA} \\
\bottomrule
\end{tabular}
\caption{
    Artifacts used in this work and their licenses and links. The number of parameters of the models are shown in parentheses.
}
\label{tab:artifacts}
\end{table*}


\subsection{Created Artifacts}

We create a new training framework in this work.
We release the code for the training framework and several models trained with the framework under the MIT license.
We only allow research use of our code and models on personal and public data, which is compatible with the original access conditions of the models and datasets.
Using the model on other individuals without authorization is unethical and strictly forbidden.

\end{document}